\documentclass[runningheads]{llncs}
\usepackage[T1]{fontenc}
\usepackage{graphicx}
\usepackage{cite}
\usepackage{multirow}
\usepackage{subcaption}
\usepackage{amsmath}
\begin{document}
\title{HiStyle: Hierarchical Style Embedding Predictor for Text-Prompt-Guided Controllable Speech Synthesis}
\titlerunning{HiStyle: Hierarchical Style Embedding Predictor for Controllable TTS}
%
\author{Ziyu Zhang \and
Hanzhao Li \and
Jingbin Hu \and
Wenhao Li \and
Lei Xie
}
\authorrunning{Ziyu Zhang et al.}

\institute{Audio, Speech and Language Processing Group (ASLP@NPU), School of Computer Science, Northwestern Polytechnical University, Xi'an, China \\
\url{http://www.npu-aslp.org/}\\
\email{ziyu\_zhang@mail.nwpu.edu.cn}}

\maketitle             

\begin{abstract}
Controllable speech synthesis refers to the precise control of speaking style by manipulating specific prosodic and paralinguistic attributes, such as gender, volume, speech rate, pitch, and pitch fluctuation. With the integration of advanced generative models, particularly large language models (LLMs) and diffusion models, controllable text-to-speech (TTS) systems have increasingly transitioned from label-based control to natural language description-based control, which is typically implemented by predicting global style embeddings from textual prompts. However, this straightforward prediction overlooks the underlying distribution of the style embeddings, which may hinder the full potential of controllable TTS systems. In this study, we use t-SNE analysis to visualize and analyze the global style embedding distribution of various mainstream TTS systems, revealing a clear hierarchical clustering pattern: embeddings first cluster by timbre and subsequently subdivide into finer clusters based on style attributes. Based on this observation, we propose HiStyle, a two-stage style embedding predictor that hierarchically predicts style embeddings conditioned on textual prompts, and further incorporate contrastive learning to help align the text and audio embedding spaces. Additionally, we propose a style annotation strategy that leverages the complementary strengths of statistical methodologies and human auditory preferences to generate more accurate and perceptually consistent textual prompts for style control. Comprehensive experiments demonstrate that when applied to the base TTS model, HiStyle achieves significantly better style controllability than alternative style embedding predicting approaches while preserving high speech quality in terms of naturalness and intelligibility. Audio samples are available at https://anonymous.4open.science/w/HiStyle-2517/.

\keywords{
Style Controllable TTS \and Style Embedding Distribution \and Two-stage Embedding Predictor \and Contrastive Learning.}
\end{abstract}

\section{Introduction}
Speech synthesis has emerged as a critical technology in digital content creation and human-computer interaction. The advent of advanced generative models, particularly LLMs~\cite{wang2023neural, betker2023better, lajszczak2024base, anastassiou2024seed} and diffusion models~\cite{vyas2023audiobox, eskimez2024e2, chen2024f5}, has driven significant breakthroughs in speech synthesis, empowering TTS systems to generate human-like speech with remarkable speech quality and prosodic naturalness. 

In recent years, there has been a growing research focus on developing style-controllable TTS systems that allow flexible control of speaking style attributes, such as speech rate, pitch, and volume. A common approach in prior work~\cite{gao2025emo, kim2021expressive, liu2024diffstyletts} is to use categorical labels, e.g. speaking rate level or emotion category, to represent and control these style attributes. However, such label-based methods are constrained by a limited set of predefined style categories extracted from the training corpus, which inherently limits the system’s ability to generalize to new speaking styles not seen during training. To address this limitation, subsequent studies~\cite{li2025styletts, ju2024naturalspeech, kim2023sc} introduced reference-audio-based style transfer, which uses acoustic representation extracted from a reference audio to guide synthesis and thus avoids constrained by predefined style categories. However, seeking suitable style reference audios that precisely match user intent remains time-consuming and often impractical.

To achieve more flexible and user-friendly style control, several studies~\cite{yang2024instructtts, guo2023prompttts,leng2023prompttts2, zhang2023promptspeaker, vyas2023audiobox} have adopted natural language descriptions for controllable speech synthesis, enabling intuitive control over diverse speaking style attributes and avoiding the limitations of label- and reference-based methods. Typically, these approaches achieve controllable synthesis by predicting a global style embedding from a text prompt. To realize this, previous studies have explored a variety of prediction strategies. Early work~\cite{chen2024generating} employed lightweight projection networks for direct text-to-style embedding mapping, while its simplistic architecture limited controllable performance. Subsequent approaches leveraged diffusion-based variational networks to synthesize style embeddings from text prompts~\cite{leng2023prompttts2}, capitalizing on diffusion models' superior ability to capture and model speech style variations, thereby achieving enhanced controllability. However, all these methods treat style embedding prediction as a single-step mapping procedure and therefore ignore the intrinsic hierarchical distribution of the style embedding space, which may limit the controllability of the synthesized speech. In the other hand, for the text prompt labeling aspect of style-controllable datasets, most prior works~\cite{vyas2023audiobox, jin2024speechcraft, wang2025spark} relied on manually set thresholds for style attribute annotation, using hand-crafted percentage-based criteria to define attribute labels. Such approaches are inherently inflexible and fail to consider the alignment between automatically assigned labels and actual human perceptual judgments.

To address these challenges, we visualize and analyze the style embedding distribution of various mainstream TTS systems, revealing that style embeddings exhibit a clear hierarchical distribution: globally grouped according to timbre and locally subdivided by specific style attributes. Building on this insight, we design a novel hierarchical two-stage style embedding predictor, named HiStyle, which first predict coarse-grained global embeddings and then predict fine-grained style embeddings. To further enhance the alignment between textual and acoustic spaces, we incorporate a contrastive learning objective during training. In addition, recognizing the limitations of fixed thresholds in style annotation, we develop a new data annotation pipeline that combines objective statistical methods with subjective human perceptual feedback, helping generate more accurate and perceptually consistent style labels. 

In summary, the main contributions of this work are as follows:
\begin{itemize}
    \item We conduct a comprehensive visualization and analysis of the style embedding spaces in mainstream TTS systems, revealing their hierarchical organization: globally clustered by timbre and further subdivided by style attributes.
    \item We propose HiStyle, a novel two-stage style embedding predictor with contrastive learning, which hierarchically predicts style embeddings conditioned on textual prompts and achieves significantly improved controllability and naturalness in speech synthesis.
    \item We develop an improved data annotation pipeline that integrates statistical analysis with human perceptual evaluation, achieving more accurate and perceptually consistent style labeling for training and evaluation.
\end{itemize}

\section{Preliminary Analysis}\label{AA}
Contemporary text-prompt controllable TTS systems~\cite{wang2018style} commonly employ global embeddings to control speech attributes (e.g., speech rate, pitch, and timbre). These embeddings are typically extracted by learnable audio encoders and subsequently predicted from prompt texts during inference. Some studies~\cite{leng2023prompttts2} have employed Principal Component Analysis (PCA)~\cite{mackiewicz1993principal} to visualize the embedding space, revealing distinct clustering patterns corresponding to speaker identity and emotional characteristics. However, the precise relationship between the embedding space and specific style attributes (e.g., speech rate and pitch patterns) remains insufficiently characterized.

\begin{figure*}[t]
    \centering
    \subfloat[\textbf{Speaker-Level Clustering}\label{fig:tsne_hierarchy_1}]{
        \includegraphics[width=0.45\textwidth, trim=0 10 0 10, clip]{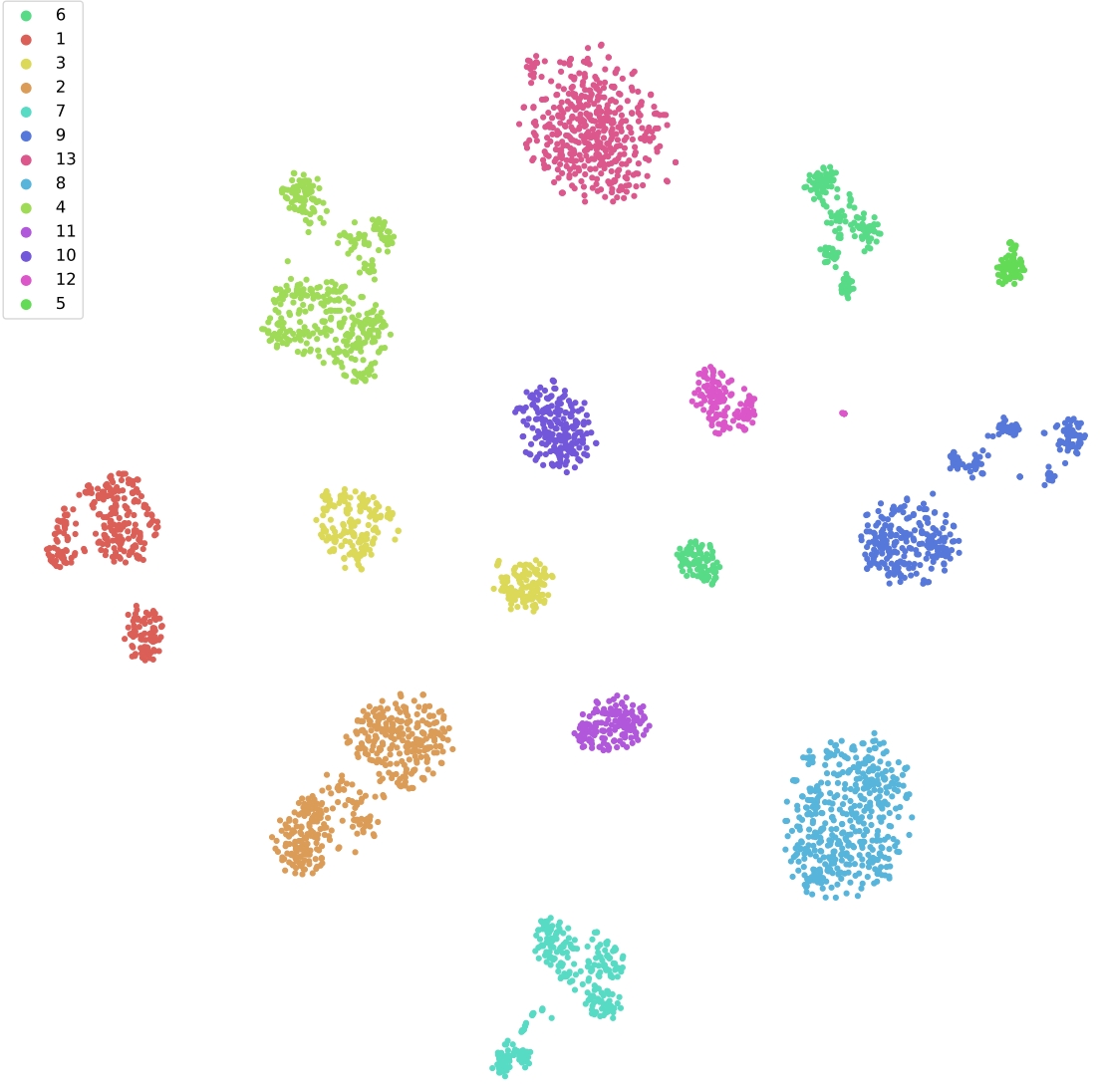}
    }
    \hspace{0.01\textwidth} 
    \subfloat[\textbf{Fluctuation-Level Clustering}\label{fig:tsne_hierarchy_2}]{
        \includegraphics[width=0.49\textwidth, trim=2 2 2 2, clip]{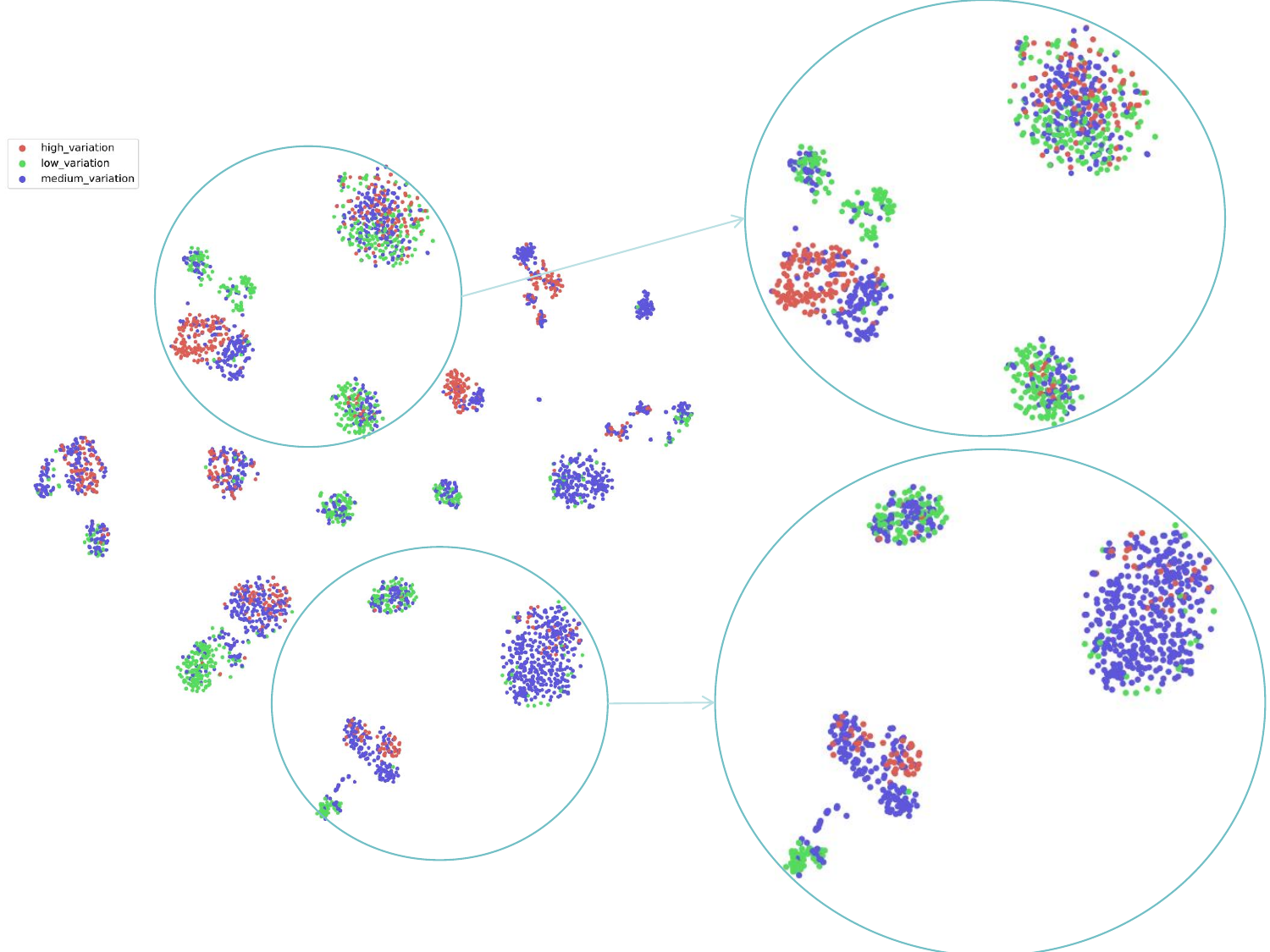}
    }
    \vspace{0.5cm} 
    \raggedright 
    \caption{t-SNE visualization of global embeddings from Transformer-based encoder: (a) Speaker-Level Clustering, (b) Pitch Fluctuation-Level Sub-Clustering.}
    \label{fig:tsne_hierarchy}
\end{figure*}

To better understand this internal structure, we select several representative reference encoder architectures and extract their global embeddings for visualization using t-SNE. We use various models to extract global embeddings for different style attributes, including the acoustic encoder based on ECAPA-TDNN~\cite{guo2024fireredtts}, the pre-trained voiceprint model~\cite{desplanques2020ecapa}, and the CNN-GRU-based encoder~\cite{guo2024npu}. To observe the distribution of different style attributes in the global embedding, we construct a highly expressive and high-quality dataset with multiple speakers and multiple speech attributes. Using this dataset, we input speech samples into each encoder to extract the corresponding global embeddings. Then, we apply t-SNE to project the high-dimensional embeddings into a 2D space. 

The Figure~\ref{fig:tsne_hierarchy} shows the visualization result in the CNN-GRU-based encoder. As shown in Figure~\ref{fig:tsne_hierarchy_1}, when we color the t-SNE visualization by speaker identity, the embedding space partitions clearly into distinct clusters corresponding to each speaker, indicating that the global embedding is organized by speaker timbre. Next, we recolor the same t-SNE cluster according to the pitch fluctuation attribute. As shown in Figure~\ref{fig:tsne_hierarchy_2}, the embeddings do not exhibit distinct grouping across the entire space. However, by comparing Figure~\ref{fig:tsne_hierarchy_1} and Figure~\ref{fig:tsne_hierarchy_2}, we find that the embeddings are further separated based on pitch fluctuation within each speaker-specific cluster. This phenomenon suggests that the embedding space has a clear hierarchical structure: embeddings are first grouped by timbre at the global level and then further organized by style attributes within each timbre cluster. We conduct the same visualization procedure using the other encoder structures and other attributes in the supplementary materials, observing the same hierarchical clustering patterns.

\section{HiStyle}

\begin{figure*}[t]
  \centering
  \includegraphics[width=1.1\textwidth]{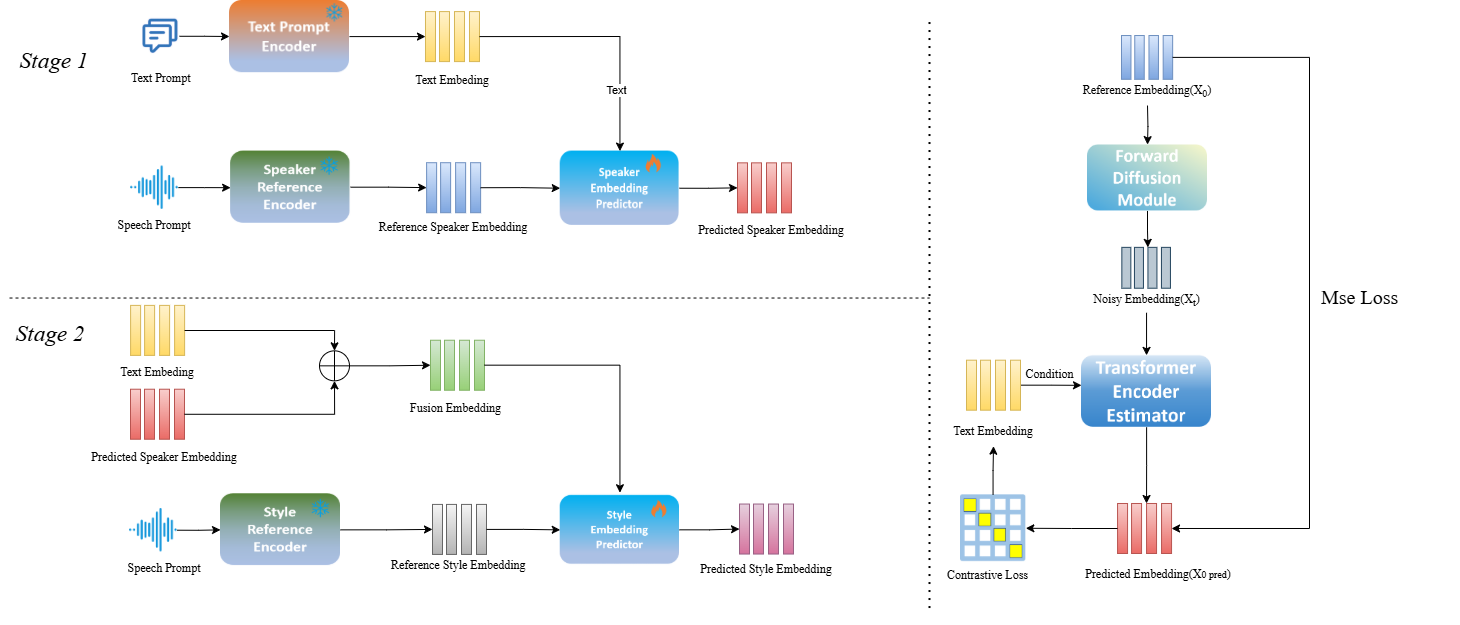}
  \caption{Overview architecture of the HiStyle embedding predictor. The subplot in the upper-left corner illustrates the prediction process of our first stage. The Speaker Embedding Predictor takes the text prompt embedding as a condition and uses the reference speaker embedding to predict the predicted speaker embedding. Similarly, the subplot in the lower-left corner represents the second stage, where the Style Embedding Predictor takes the text prompt embedding along with the residual connection of the intermediate result (predicted speaker embedding) from the first stage as conditions, and leverages the fusion embedding to predict the predicted style embedding. The subplot on the right depicts the detailed architecture and training process of the two Embedding Predictors.
}
  \label{fig:twostage_framework}
\end{figure*}


Inspired by the hierarchical structure observed in the previous section, we propose a novel hierarchical two-stage embedding predictor, as illustrated in Figure~\ref{fig:twostage_framework}. Our predictor consists of three key components: the text prompt encoder, speaker embedding predictor and style embedding predictor. First, the text prompt encoder comprising a pre-trained BERT model followed by a linear projection encodes the input text description into a text-prompt embedding\cite{devlin2019bert}. Next, we implement a speaker embedding predictor that predicts a global speaker-related embedding conditioned on the text-prompt embedding. This predicted embedding captures the speaker-related information comprising both timbre and style information and serves as an intermediate representation\cite{wang2018style, chen2022fine}. Then, we employ a style embedding predictor, which fuse the predicted speaker embedding with the original text-prompt embedding through a residual connection as condition to predict the final style embedding\cite{he2016deep}. 

We adopt a unified architecture for \textit{Speaker Embedding Predictor} and \textit{Style Embedding Predictor}, all implemented within a conditional diffusion model based on transformer blocks~\cite{vaswani2017attention}. As illustrated in Figure~\ref{fig:twostage_framework}, this model structure consists of a forward diffusion module for noise-adding  and a transformer encoder-based estimator module for denoising\cite{liu2022diffsinger}.

During training, the diffusion model adds Gaussian noise to the ground-truth reference embedding $\mathbf{x}_0$, resulting in a noisy feature $\mathbf{x}_t$. This can be formulated as
\begin{equation}
    \mathbf{x}_t = \sqrt{\bar{\alpha}_t}\,\mathbf{x}_0 + \sqrt{1-\bar{\alpha}_t}\,\mathbf{z}, \quad \mathbf{z} \sim \mathcal{N}(\mathbf{0}, \mathbf{I}),
\end{equation}
where $\bar{\alpha}_t$ denotes the noise schedule and $\mathbf{z}$ is standard Gaussian noise. 

The global noisy reference embedding $\mathbf{x}_t$, together with the text embedding and the current diffusion step, are concatenated and fed into the transformer blocks. This architecture fuses all three sources of information via multi-layer self-attention, yielding a global contextual representation. Then the model predicts the reference embedding $\mathbf{x}_{0\_\text{pred}}$ conditioned on this global contextual representation.


The model is trained to minimize the Mean Squared Error (MSE) loss between its prediction and the ground-truth reference embedding:
\begin{equation}
    \mathcal{L}_\mathrm{MSE} = \left\| \mathbf{x}_{0\_\text{pred}} - \mathbf{x}_0 \right\|_2^2.
\end{equation}

To further enhance the alignment between the textual and acoustic embedding spaces, we additionally incorporate a contrastive learning objective based on cosine similarity loss\cite{radford2021learning,oord2018representation}. For each training batch, we construct positive pairs by matching each predicted embedding with its corresponding text prompt embedding, and negative pairs by associating it with the text prompt embeddings of other samples within the same batch. 

We define the cosine similarity loss as:
\begin{equation}
\mathcal{L}_{\text{CL}} = 1 - \cos(\mathbf{z}_\text{pred}, \mathbf{z}_\text{ref}),
\end{equation}
where $\mathbf{z}_\text{pred}$ is the predicted embedding and $\mathbf{z}_\text{ref}$ is the corresponding reference embedding. For negative samples $\mathbf{z}_\text{neg}$, we optionally include a repulsion term:
\begin{equation}
\mathcal{L}_{\text{neg}} = \sum{i=1}^{N-1} \max\left(0, \cos(\mathbf{z}_\text{pred}, \mathbf{z}_{\text{neg},i}) - m\right),
\end{equation}
where $m$ is a margin that pushes negative pairs to have similarity below a threshold.

The final contrastive loss combines positive and negative terms:
\begin{equation}
\mathcal{L}_{\text{contrastive}} = \mathcal{L}_{\text{CL}} + \lambda_{\text{neg}} \cdot \mathcal{L}_{\text{neg}}.
\end{equation}

The overall training objective is:
\begin{equation}
\mathcal{L}_{\text{total}} = \mathcal{L}_{\text{MSE}} + \lambda \cdot \mathcal{L}_{\text{contrastive}},
\end{equation}
where $\lambda$ and $\lambda_{\text{neg}}$ are hyperparameters that balance the contribution of the contrastive terms.

During inference, the model starts from a random Gaussian noised embedding and progressively denoise it using only the text embedding as condition, ultimately generating a predicted reference style embedding that aligns with the input text prompt.

Above we introduce the two-stage prediction approach and the specific structure of each predictor, and below we will introduce how our predictor can be applied to the TTS model for controllable speech synthesis. Our proposed style embedding predictor is highly generalizable and can be flexibly integrated into mainstream TTS systems. The core idea is to design a speaker reference encoder and a style reference encoder within the TTS model to extract corresponding global speaker-related and style embeddings. These two embeddings are then fed into our predictor, which hierarchically predicts the target speaker embedding and target style embedding conditioned on the text prompt. In practical implementation, the speaker reference encoder can be flexibly chosen from various architectures, such as an ECAPA-TDNN model or a pre-trained voiceprint model. While the style reference encoder, on the other hand, can adopt architectures like transformer-based models that are better suited for capturing complex stylistic attributes.

\section{Data Annotation}
We leverage a high-quality internal dataset containing 2,000 hours of expressive speech with over 20 distinct timbres and diverse styles to provide rich stylistic variations. Our annotation pipeline integrates a dual approach combining statistical analysis with human perceptual evaluation, achieving both precise and perceptually consistent style labeling. Our annotation pipeline consequently follows three key steps:

\subsection{Attribute Value Computation} 

The first step involves quantifying speech style attributes into measurable parameters, where we calculate gender, speech rate, volume, pitch, and pitch fluctuation values\cite{lyth2024natural}.

For speech rate, we first use signal processing methods to trim the silent segments at the beginning and end of the speech, then extract the phoneme sequence and audio duration with internal tools. The relative speech rate is defined as the total number of phonemes divided by the total duration. In our analysis, we observe a significant difference in calculated speech rate values between Chinese and English utterances, even when the perceived speaking speed is similar. As shown in Table~\ref{tab:lang_rate_diff}, the average speech rate for English is notably higher than that for Chinese, primarily due to differences in phoneme length between the two languages. To ensure annotation accuracy, we therefore set separate grading thresholds for Chinese and English speech.


\begin{table}[h]
\centering
\caption{Mean and standard deviation of computed speech rate for Chinese and English utterances.}
\setlength{\tabcolsep}{10pt} 
\renewcommand{\arraystretch}{1.5} 
\begin{tabular}{lcc}
\hline
Language & Mean\_value(s) & Std\_value(s) \\
\hline
Chinese  & 14.63 & 4.94 \\
English  & 18.25 & 7.70 \\
\hline
\end{tabular}
\label{tab:lang_rate_diff}
\end{table}

For gender, we fine-tune the ECAPA-TDNN\cite{desplanques2020ecapa} model on a large internal gender-labeled dataset to ensure robust generalization, then use its output probabilities for annotation. For pitch and its variability (pitch fluctuation), we use \textsc{PyWorld} to extract the frame-level fundamental frequency ($F_0$) contour for each audio. To ensure accurate measurements, we remove abnormal zeros in the $F_0$ array caused by unnatural pauses or noise. We then calculate the mean and standard deviation of the remaining non-zero $F_0$ values, using them as the pitch and fluctuation of each recording, respectively.

\subsection{Statistic-based Classification} 

The second step aims to assign each style attribute to an appropriate category, according to the precise values calculated in the first step. The specific procedure is as follows:

For gender annotation, we assign the category with the highest probability as the final gender label for each utterance. For the other attributes, we adopt a strategy that combines objective statistical thresholds with subjective auditory preference. Specifically, for speech rate, pitch, and pitch fluctuation, we use a three-level classification scheme. For each attribute, we compute the mean and standard deviation across the relevant data groups, calculating separately by gender for pitch and fluctuation, and separately by language for speech rate. Initial thresholds are then set using $\mu-\sigma$ and $\mu+\sigma$ as boundaries, where $\mu$ and $\sigma$ are the mean and standard deviation for each group, providing a sound statistical foundation for grading. 

\subsection{Human Perception Adjustment}
\begin{figure}[t] 
    \centering
    \includegraphics[width=0.8\linewidth , trim=0 20 0 100, clip]{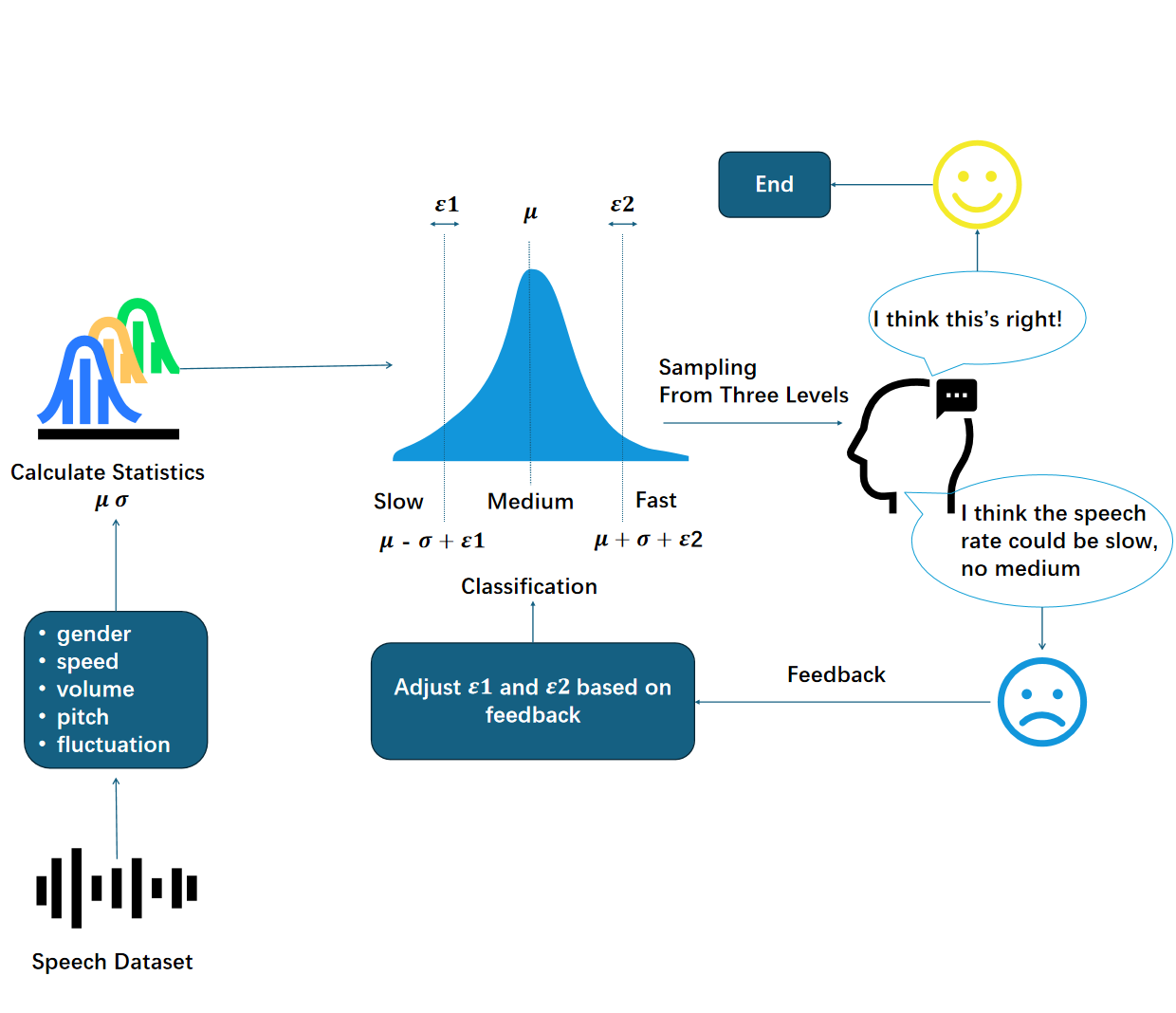}
    \caption{Iterative Annotation Pipeline Combining Statistical Thresholding and Human Perception}
    \label{fig:data_pipeline}
\end{figure}
To further align these calculated thresholds with human perception, we design an iterative annotation workflow, as illustrated in Figure~\ref{fig:data_pipeline}. After using the initial thresholds to categorize utterances, we randomly sample a set of borderline cases. Specifically, samples falling within a ±5\% margin of each threshold boundary for perceptual evaluation. In each iteration, we invite three annotators with basic speech knowledge to independently listen to 50 utterances per attribute (randomly selected and distributed across levels) and assign perceptual labels such as “slow,” “medium,” or “fast” for speech rate. After collecting the annotations, we analyze inter-annotator agreement and compare the perceptual labels with the current threshold-based labels. If consistent discrepancies are observed (for example, if more than 80\% of listeners consistently classify a threshold-defined “medium” utterance as “fast”), we adjust the corresponding threshold accordingly. This process is repeated for 2–3 rounds until the classification results achieve over 85\% agreement with human perceptual judgments, ensuring that the final attribute intervals are both statistically grounded and perceptually reliable.

Finally, we combine all the attribute-level descriptions using a large language model\footnote{We use ChatGPT (gpt-3.5-turbo) as the LLM.} to generate fluent and contextually appropriate natural language sentences. This approach ensures that the resulting style descriptions are cohesive, richly detailed, and exhibit human-like expressiveness.

\section{Experiment}\label{sec:typestyle}

\subsection{Training Setup and Model Details}

\textbf{Datasets} We utilize a high-quality internal dataset comprising 2,000 hours of expressive speech, spanning over 20 distinct timbres and a wide range of style attribute levels. Using the annotation pipeline described above, we label each utterance with a level of each style attribute and combine these annotations into a descriptive sentence to serve as the text prompt. And then we preserve 2,000 utterances as the test set and use the remaining data for training.

\textbf{Model and Training Details} The diffusion encoder in our system, including the speaker embedding predictor, and style embedding predictor consisting of 12 layers, each with a hidden size of 512. Each diffusion model contains approximately 30 million parameters. We train our two-stage embedding predictor using the Adam optimizer with a learning rate of $2 \times 10^{-4}$, employing a warmup schedule followed by cosine decay. The batch size is set to 128, and training is conducted on 8 NVIDIA A6000 GPUs.

\subsection{Evaluation Metrics}

\textbf{Objective Metrics} We first assess the accuracy of key speaking style attributes, including gender, speech rate, volume, pitch, and fluctuation. Gender accuracy is determined using an internal fine-tuned gender classifier. For the remaining attributes, we utilize the style annotation pipeline to annotate the synthesized speech, and then compare these labels with the corresponding style attributes in the text prompt to calculate the accuracy for each attribute. Additionally, we evaluate the overall quality of the synthesized speech in terms of Word Error Rate (WER), and UTMOS\cite{saeki2022utmos}, following the evaluation protocols provided by Seed-TTS-eval\cite{anastassiou2024seed}.

\textbf{Subjective Metrics} We use the Mean Opinion Score (MOS) to evaluate speech naturalness (N-MOS) and style consistency (Style-MOS).

\subsection{Effectiveness of HiStyle}
In this section, we present a series of objective and subjective experiments to evaluate the effectiveness of various style embedding prediction strategies for speech style control. All methods are implemented on a unified backbone: a SingleCodec-based TTS system that utilizes a LLaMA\cite{touvron2023llama} architecture as its language model. This backbone is chosen for its simplicity and strong baseline performance. We compare our method with five different approaches:

\textbf{Text Prompt Only} Following the implementation in PromptTTS\cite{guo2023prompttts}, we directly use the text-prompt embedding extracted by the prompt encoder as the style embedding, without any audio supervision. The prediction network is optimized only by speech synthesis loss.

\textbf{Discriminative model} Following the implementation in \cite{chen2024generating}, we train a lightweight projection module to map the text-audio embedding space, taking the projected vector as the style embedding. The prediction network is optimized by discriminative loss and speech synthesis loss.

\textbf{Variation Network} Following the implementation in PromptTTS2\cite{leng2023prompttts2}, we employ a generative variation network to generate a reference-audio embedding conditioned on the text-prompt embedding, using this synthesized embedding as the style embedding. The prediction network is optimized by generative loss and speech synthesis loss.

\textbf{Query Encoder} Following the implementation in FleSpeech\cite{li2025flespeech}, we use the query encoder and diffusion network to map the text-to-audio embedding space, the prediction network is optimized by diffusion loss and speech synthesis loss.

\textbf{HiStyle} Our proposed method, which using two transformer encoder based diffusion models to hierarchically predict style embeddings conditioned on text prompts. The prediction network is optimized by MSE loss, contrastive loss and speech synthesis loss.

       



        



\begin{table*}[htbp]
\centering
\caption{Performance of Different Style Embedding Prediction Methods on Objective and Subjective Metrics}
\resizebox{\textwidth}{!}{
\begin{tabular}{c|ccccc|cc|cc}
\hline
\multirow{2}{*}{\textbf{Model}} & \multicolumn{5}{c|}{\textbf{Accuracy(\%)↑}} & \multirow{2}{*}{\textbf{WER(\%)↓}} & \multirow{2}{*}{\textbf{UTMOS↑}} & \multirow{2}{*}{\textbf{N-MOS↑}} & \multirow{2}{*}{\textbf{Style-MOS↑}} \\ \cline{2-6}
                               & \textbf{Gender} & \textbf{Speed} & \textbf{Volume} & \textbf{Pitch} & \textbf{Fluctuation} & & & & \\ \hline

\textbf{Text Prompt Only}       & \underline{98.75}  & 89.21  & 85.32  & 85.69  & 82.32  & \textbf{3.09}  & 3.37  & 3.79 ± 0.03 & 3.45 ± 0.04 \\ 

\textbf{Discriminative Model}   & 97.71  & 85.21  & 88.43  & 90.65  & 83.65  & 3.69  & 3.36  & 3.71 ± 0.08 & 3.48 ± 0.02 \\ 

\textbf{Variation Network}      & 98.27  & \textbf{92.56} & \underline{93.33} & 88.21  & 86.58  & 4.29  & 3.38  & 3.69 ± 0.06 & 3.52 ± 0.05 \\ 

\textbf{Query Encoder}          & 97.66  & 90.48  & 91.58  & \underline{91.86} & 83.31  & 3.82  & \textbf{3.45}  & \textbf{3.83 ± 0.09} & 3.68 ± 0.03 \\ 

\textbf{HiStyle}               & \textbf{98.88}  & \underline{90.98} & \textbf{95.56} & \textbf{92.87} & \textbf{88.02} & \underline{3.32} & \underline{3.41} & \underline{3.80 ± 0.08} & \textbf{3.71 ± 0.05} \\ \hline
\end{tabular}
}
\label{tab:obj}
\end{table*}

As shown in Table~\ref{tab:obj}, our proposed two-stage embedding predictor consistently outperforms all compared approaches across multiple objective metrics. In terms of style attribute controllability, our method achieves the highest accuracy across key dimensions including volume (95.56\%), pitch (92.87\%), and fluctuation (88.02\%), indicating its superior capability in capturing diverse speaking style characteristics from text prompts. In addition, it achieves the highest accuracy in gender prediction (98.88\%), confirming the robustness of the model in handling both linguistic and paralinguistic cues. Moreover, our method maintains a low WER of 3.32\% and a low UTMOS of 3.41, demonstrating that enhanced controllability does not come at the cost of speech intelligibility or naturalness. Additionally, for subjective indicators, our method achieves a Style-MOS score of 3.71 ± 0.05, the highest among all competing systems, demonstrating superior style controllability and a stronger alignment between the intended and perceived speaking style. In terms of naturalness, our method obtains a N-MOS of 3.80 ± 0.08, which is competitive with the best performing baseline (Query Encoder: 3.83 ± 0.09). Notably, our approach delivers the best overall trade-off between style controllability and speech naturalness.

In contrast, the Text Prompt Only approach shows the lowest performance in all style dimensions except gender. This suggests that using text prompts alone, without leveraging reference audio embeddings, is insufficient for modeling the inherent variability in speaking styles, resulting in poor controllability. The Discriminative Model delivers modest performance, likely due to its overly simplistic projection architecture, which limits its capacity to capture complex style representations. The Variation Network demonstrates strong performance in speed accuracy (92.56\%) but suffers from the highest WER (4.29\%), revealing a clear trade-off between style precision and linguistic consistency. Meanwhile, the Query Encoder yields more balanced results across style dimensions but still falls short of our model in both accuracy and perceptual quality. These comparisons highlight the advantage of our hierarchical predicting strategy, which achieves both controllable and intelligible speech synthesis.

\subsection{Ablation Study}
\begin{table*}[htbp]
\centering
\caption{Ablation Study on Contrastive Learning and Style Annotation Strategy}
\resizebox{\textwidth}{!}{
\begin{tabular}{c|ccccc|cc|cc}
\hline
\multirow{2}{*}{Model} & \multicolumn{5}{c|}{Accuracy(\%)↑} & \multirow{2}{*}{WER(\%)↓} & \multirow{2}{*}{UTMOS↑} & \multirow{2}{*}{N-MOS↑} & \multirow{2}{*}{Style-MOS↑} \\
                        & Gender & Speed & Volume & Pitch & Fluctuation & & & &  \\ \hline

\textbf{HiStyle}       & \textbf{98.78} & \textbf{90.84} & \textbf{95.43} & \textbf{91.97} & \textbf{88.72} & \textbf{3.25} & \textbf{3.39} & \textbf{3.86 ± 0.07} & \textbf{3.75 ± 0.03} \\

w/o Contrastive Learning  & \underline{94.49} & \underline{88.10} & \underline{90.43} & \underline{89.78} & \underline{83.67} & 3.96 & \underline{3.29} & 3.78 ± 0.01 & 3.66 ± 0.06 \\

w/o Style Annotation    & 92.64 & 85.98 & 88.67 & 80.02 & 82.21 & \underline{3.58} & 3.20 & \underline{3.84 ± 0.06} & \underline{3.68 ± 0.08} \\

w/o Both                & 92.41 & 85.48 & 88.49 & 83.88 & 80.82 & 4.08 & 3.08 & 3.69 ± 0.04 & 3.62 ± 0.02 \\

\hline
\end{tabular}
}
\label{tab:ablation}
\end{table*}


Table~\ref{tab:ablation} presents the ablation results of HiStyle across four configurations (full model, without contrastive learning, without style annotation, and without both). For objective evaluation, the full model (HiStyle) achieves the best performance, excelling across all style dimensions, particularly in gender (98.78\%), speech rate (90.84\%), and volume (95.43\%). Removing contrastive learning results in a decline in all accuracy metrics, with gender accuracy dropping to 94.49\%. While omitting style annotation leads to a significant decrease in style control accuracy (e.g., pitch drops to 80.02\%), although the WER remains at a moderate 3.58\%. Removing both contrastive learning and style annotation results in the worst performance, with a significant increase in WER to 4.08\%, emphasizing the complementarity and importance of contrastive learning and style annotation.

In terms of subjective evaluation, HiStyle also shows excellent performance in N-MOS and Style-MOS. The full model achieves the highest scores in both N-MOS (3.86 ± 0.07) and Style-MOS (3.75 ± 0.03) among all configurations, indicating that HiStyle excels not only in speech naturalness but also in maintaining style consistency. Removing contrastive learning causes a slight decline in both N-MOS and Style-MOS, with scores of 3.78 ± 0.01 and 3.66 ± 0.06, respectively. Omitting style annotation has a more significant impact, particularly on Style-MOS, which drops to 3.68 ± 0.08. The configuration without both contrastive learning and style annotation further reduces these metrics to 3.69 ± 0.04 for N-MOS and 3.62 ± 0.02 for Style-MOS, highlighting the importance of contrastive learning and style annotation in improving both speech naturalness and style consistency.

\section{Conclusions}
In this study, we introduce HiStyle, a hierarchical two-stage style embedding predictor for text-prompt-guided controllable speech synthesis. Through an in-depth analysis of the global style embedding distribution in text-to-speech systems, we identify a clear hierarchical structure, with embeddings initially clustered by timbre and further subdivided by specific style attributes. Leveraging this insight, HiStyle is designed to predict both coarse-grained speaker embeddings and fine-grained style embeddings. To further enhance alignment between textual prompts and acoustic representations, we incorporate contrastive learning, strengthening the connection between the text and audio embedding spaces. Additionally, our novel style annotation strategy, combining statistical methodologies and human perceptual evaluation, facilitated the creation of accurate, perceptually consistent text prompts for style control. Experimental results demonstrate that HiStyle outperforms existing style embedding prediction methods in both objective and subjective metrics, achieving superior style controllability and maintaining high naturalness and intelligibility in the generated speech. The proposed approach paves the way for more versatile and user-friendly style-controllable speech synthesis systems.


\bibliographystyle{splncs04}
\bibliography{ref}

\begin{thebibliography}{10}

\bibitem{wang2023neural}
Chengyi Wang, Sanyuan Chen, Yu~Wu, Ziqiang Zhang, Long Zhou, Shujie Liu, Zhuo Chen, Yanqing Liu, Huaming Wang, Jinyu Li, et~al.,
\newblock ``Neural codec language models are zero-shot text to speech synthesizers,''
\newblock {\em arXiv preprint arXiv:2301.02111}, 2023.

\bibitem{betker2023better}
James Betker,
\newblock ``Better speech synthesis through scaling,''
\newblock {\em arXiv preprint arXiv:2305.07243}, 2023.

\bibitem{lajszczak2024base}
Mateusz {\L}ajszczak, Guillermo C{\'a}mbara, Yang Li, Fatih Beyhan, Arent Van~Korlaar, Fan Yang, Arnaud Joly, {\'A}lvaro Mart{\'\i}n-Cortinas, Ammar Abbas, Adam Michalski, et~al.,
\newblock ``Base tts: Lessons from building a billion-parameter text-to-speech model on 100k hours of data,''
\newblock {\em arXiv preprint arXiv:2402.08093}, 2024.

\bibitem{anastassiou2024seed}
Philip Anastassiou, Jiawei Chen, Jitong Chen, Yuanzhe Chen, Zhuo Chen, Ziyi Chen, Jian Cong, Lelai Deng, Chuang Ding, Lu~Gao, et~al.,
\newblock ``Seed-tts: A family of high-quality versatile speech generation models,''
\newblock {\em arXiv preprint arXiv:2406.02430}, 2024.

\bibitem{vyas2023audiobox}
Apoorv Vyas, Bowen Shi, Matthew Le, Andros Tjandra, Yi-Chiao Wu, Baishan Guo, Jiemin Zhang, Xinyue Zhang, Robert Adkins, William Ngan, et~al.,
\newblock ``Audiobox: Unified audio generation with natural language prompts,''
\newblock {\em arXiv preprint arXiv:2312.15821}, 2023.

\bibitem{eskimez2024e2}
Sefik~Emre Eskimez, Xiaofei Wang, Manthan Thakker, Canrun Li, Chung-Hsien Tsai, Zhen Xiao, Hemin Yang, Zirun Zhu, Min Tang, Xu~Tan, et~al.,
\newblock ``E2 tts: Embarrassingly easy fully non-autoregressive zero-shot tts,''
\newblock in {\em 2024 IEEE Spoken Language Technology Workshop (SLT)}. IEEE, 2024, pp. 682--689.

\bibitem{chen2024f5}
Yushen Chen, Zhikang Niu, Ziyang Ma, Keqi Deng, Chunhui Wang, Jian Zhao, Kai Yu, and Xie Chen,
\newblock ``F5-tts: A fairytaler that fakes fluent and faithful speech with flow matching,''
\newblock {\em arXiv preprint arXiv:2410.06885}, 2024.

\bibitem{gao2025emo}
Xiaoxue Gao, Chen Zhang, Yiming Chen, Huayun Zhang, and Nancy~F Chen,
\newblock ``Emo-dpo: Controllable emotional speech synthesis through direct preference optimization,''
\newblock in {\em ICASSP 2025-2025 IEEE International Conference on Acoustics, Speech and Signal Processing (ICASSP)}. IEEE, 2025, pp. 1--5.

\bibitem{kim2021expressive}
Minchan Kim, Sung~Jun Cheon, Byoung~Jin Choi, Jong~Jin Kim, and Nam~Soo Kim,
\newblock ``Expressive text-to-speech using style tag,''
\newblock {\em arXiv preprint arXiv:2104.00436}, 2021.

\bibitem{liu2024diffstyletts}
Jiaxuan Liu, Zhaoci Liu, Yajun Hu, Yingying Gao, Shilei Zhang, and Zhenhua Ling,
\newblock ``Diffstyletts: Diffusion-based hierarchical prosody modeling for text-to-speech with diverse and controllable styles,''
\newblock {\em arXiv preprint arXiv:2412.03388}, 2024.

\bibitem{li2025styletts}
Yinghao~Aaron Li, Cong Han, and Nima Mesgarani,
\newblock ``Styletts: A style-based generative model for natural and diverse text-to-speech synthesis,''
\newblock {\em IEEE Journal of Selected Topics in Signal Processing}, 2025.

\bibitem{ju2024naturalspeech}
Zeqian Ju, Yuancheng Wang, Kai Shen, Xu~Tan, Detai Xin, Dongchao Yang, Yanqing Liu, Yichong Leng, Kaitao Song, Siliang Tang, et~al.,
\newblock ``Naturalspeech 3: Zero-shot speech synthesis with factorized codec and diffusion models,''
\newblock {\em arXiv preprint arXiv:2403.03100}, 2024.

\bibitem{kim2023sc}
Daegyeom Kim, Seongho Hong, and Yong-Hoon Choi,
\newblock ``Sc vall-e: Style-controllable zero-shot text to speech synthesizer,''
\newblock {\em arXiv preprint arXiv:2307.10550}, 2023.

\bibitem{yang2024instructtts}
Dongchao Yang, Songxiang Liu, Rongjie Huang, Chao Weng, and Helen Meng,
\newblock ``Instructtts: Modelling expressive tts in discrete latent space with natural language style prompt,''
\newblock {\em IEEE/ACM Transactions on Audio, Speech, and Language Processing}, 2024.

\bibitem{guo2023prompttts}
Zhifang Guo, Yichong Leng, Yihan Wu, Sheng Zhao, and Xu~Tan,
\newblock ``Prompttts: Controllable text-to-speech with text descriptions,''
\newblock in {\em ICASSP 2023-2023 IEEE International Conference on Acoustics, Speech and Signal Processing (ICASSP)}. IEEE, 2023, pp. 1--5.

\bibitem{leng2023prompttts2}
Yichong Leng, Zhifang Guo, Kai Shen, Xu~Tan, Zeqian Ju, Yanqing Liu, Yufei Liu, Dongchao Yang, Leying Zhang, Kaitao Song, et~al.,
\newblock ``Prompttts 2: Describing and generating voices with text prompt,''
\newblock {\em arXiv preprint arXiv:2309.02285}, 2023.

\bibitem{zhang2023promptspeaker}
Yongmao Zhang, Guanghou Liu, Yi~Lei, Yunlin Chen, Hao Yin, Lei Xie, and Zhifei Li,
\newblock ``Promptspeaker: Speaker generation based on text descriptions,''
\newblock in {\em 2023 IEEE Automatic Speech Recognition and Understanding Workshop (ASRU)}. IEEE, 2023, pp. 1--7.

\bibitem{chen2024generating}
Zhengyang Chen, Xuechen Liu, Erica Cooper, Junichi Yamagishi, and Yanmin Qian,
\newblock ``Generating speakers by prompting listener impressions for pre-trained multi-speaker text-to-speech systems,''
\newblock {\em arXiv preprint arXiv:2406.08812}, 2024.

\bibitem{jin2024speechcraft}
Zeyu Jin, Jia Jia, Qixin Wang, Kehan Li, Shuoyi Zhou, Songtao Zhou, Xiaoyu Qin, and Zhiyong Wu,
\newblock ``Speechcraft: A fine-grained expressive speech dataset with natural language description,''
\newblock in {\em Proceedings of the 32nd ACM International Conference on Multimedia}, 2024, pp. 1255--1264.

\bibitem{wang2025spark}
Xinsheng Wang, Mingqi Jiang, Ziyang Ma, Ziyu Zhang, Songxiang Liu, Linqin Li, Zheng Liang, Qixi Zheng, Rui Wang, Xiaoqin Feng, et~al.,
\newblock ``Spark-tts: An efficient llm-based text-to-speech model with single-stream decoupled speech tokens,''
\newblock {\em arXiv preprint arXiv:2503.01710}, 2025.

\bibitem{wang2018style}
Yuxuan Wang, Daisy Stanton, Yu~Zhang, RJ-Skerry Ryan, Eric Battenberg, Joel Shor, Ying Xiao, Ye~Jia, Fei Ren, and Rif~A Saurous,
\newblock ``Style tokens: Unsupervised style modeling, control and transfer in end-to-end speech synthesis,''
\newblock in {\em International conference on machine learning}. PMLR, 2018, pp. 5180--5189.

\bibitem{mackiewicz1993principal}
Andrzej Ma{\'c}kiewicz and Waldemar Ratajczak,
\newblock ``Principal components analysis (pca),''
\newblock {\em Computers \& Geosciences}, vol. 19, no. 3, pp. 303--342, 1993.

\bibitem{guo2024fireredtts}
Hao-Han Guo, Yao Hu, Kun Liu, Fei-Yu Shen, Xu~Tang, Yi-Chen Wu, Feng-Long Xie, Kun Xie, and Kai-Tuo Xu,
\newblock ``Fireredtts: A foundation text-to-speech framework for industry-level generative speech applications,''
\newblock {\em arXiv preprint arXiv:2409.03283}, 2024.

\bibitem{desplanques2020ecapa}
Brecht Desplanques, Jenthe Thienpondt, and Kris Demuynck,
\newblock ``Ecapa-tdnn: Emphasized channel attention, propagation and aggregation in tdnn based speaker verification,''
\newblock {\em arXiv preprint arXiv:2005.07143}, 2020.

\bibitem{guo2024npu}
Dake Guo, Jixun Yao, Xinfa Zhu, Kangxiang Xia, Zhao Guo, Ziyu Zhang, Yao Wang, Jie Liu, and Lei Xie,
\newblock ``The npu-hwc system for the iscslp 2024 inspirational and convincing audio generation challenge,''
\newblock in {\em 2024 IEEE 14th International Symposium on Chinese Spoken Language Processing (ISCSLP)}. IEEE, 2024, pp. 616--620.

\bibitem{devlin2019bert}
Jacob Devlin, Ming-Wei Chang, Kenton Lee, and Kristina Toutanova,
\newblock ``Bert: Pre-training of deep bidirectional transformers for language understanding,''
\newblock in {\em Proceedings of the 2019 conference of the North American chapter of the association for computational linguistics: human language technologies, volume 1 (long and short papers)}, 2019, pp. 4171--4186.

\bibitem{chen2022fine}
Li-Wei Chen and Alexander Rudnicky,
\newblock ``Fine-grained style control in transformer-based text-to-speech synthesis,''
\newblock in {\em ICASSP 2022-2022 IEEE International Conference on Acoustics, Speech and Signal Processing (ICASSP)}. IEEE, 2022, pp. 7907--7911.

\bibitem{he2016deep}
Kaiming He, Xiangyu Zhang, Shaoqing Ren, and Jian Sun,
\newblock ``Deep residual learning for image recognition,''
\newblock in {\em Proceedings of the IEEE conference on computer vision and pattern recognition}, 2016, pp. 770--778.

\bibitem{vaswani2017attention}
Ashish Vaswani, Noam Shazeer, Niki Parmar, Jakob Uszkoreit, Llion Jones, Aidan~N Gomez, {\L}ukasz Kaiser, and Illia Polosukhin,
\newblock ``Attention is all you need,''
\newblock {\em Advances in neural information processing systems}, vol. 30, 2017.

\bibitem{liu2022diffsinger}
Jinglin Liu, Chengxi Li, Yi~Ren, Feiyang Chen, and Zhou Zhao,
\newblock ``Diffsinger: Singing voice synthesis via shallow diffusion mechanism,''
\newblock in {\em Proceedings of the AAAI conference on artificial intelligence}, 2022, vol.~36, pp. 11020--11028.

\bibitem{radford2021learning}
Alec Radford, Jong~Wook Kim, Chris Hallacy, Aditya Ramesh, Gabriel Goh, Sandhini Agarwal, Girish Sastry, Amanda Askell, Pamela Mishkin, Jack Clark, et~al.,
\newblock ``Learning transferable visual models from natural language supervision,''
\newblock in {\em International conference on machine learning}. PmLR, 2021, pp. 8748--8763.

\bibitem{oord2018representation}
Aaron van~den Oord, Yazhe Li, and Oriol Vinyals,
\newblock ``Representation learning with contrastive predictive coding,''
\newblock {\em arXiv preprint arXiv:1807.03748}, 2018.

\bibitem{lyth2024natural}
Dan Lyth and Simon King,
\newblock ``Natural language guidance of high-fidelity text-to-speech with synthetic annotations,''
\newblock {\em arXiv preprint arXiv:2402.01912}, 2024.

\bibitem{saeki2022utmos}
Takaaki Saeki, Detai Xin, Wataru Nakata, Tomoki Koriyama, Shinnosuke Takamichi, and Hiroshi Saruwatari,
\newblock ``Utmos: Utokyo-sarulab system for voicemos challenge 2022,''
\newblock {\em arXiv preprint arXiv:2204.02152}, 2022.

\bibitem{touvron2023llama}
Hugo Touvron, Thibaut Lavril, Gautier Izacard, Xavier Martinet, Marie-Anne Lachaux, Timoth{\'e}e Lacroix, Baptiste Rozi{\`e}re, Naman Goyal, Eric Hambro, Faisal Azhar, et~al.,
\newblock ``Llama: Open and efficient foundation language models,''
\newblock {\em arXiv preprint arXiv:2302.13971}, 2023.

\bibitem{li2025flespeech}
Hanzhao Li, Yuke Li, Xinsheng Wang, Jingbin Hu, Qicong Xie, Shan Yang, and Lei Xie,
\newblock ``Flespeech: Flexibly controllable speech generation with various prompts,''
\newblock {\em arXiv preprint arXiv:2501.04644}, 2025.

\end{thebibliography}

\end{document}